\pdfoutput=1

\documentclass[11pt]{article}

\usepackage[]{ACL2023}

\usepackage{times}
\usepackage{latexsym}

\usepackage[T1]{fontenc}

\usepackage{graphicx}
\usepackage{adjustbox}
\usepackage[utf8]{inputenc}
\usepackage[english,ukrainian]{babel}
\usepackage{tempora}

\usepackage{microtype}

\usepackage{inconsolata}

\title{When a Language Question Is at Stake.

A Revisited Approach to Label Sensitive Content}

\author{Daria  Stetsenko\\
  NASK National Research Institute \\ Warsaw, Poland\\
  \texttt{daria.stetsenko@nask.pl}}

\begin{document}
\maketitle
\begin{abstract}

Many under-resourced languages require high-quality datasets for specific tasks such as offensive language detection, disinformation, or misinformation identification. However, the intricacies of the content may have a detrimental effect on the annotators. The article aims to revisit an approach of pseudo-labeling sensitive data on the example of Ukrainian tweets covering the Russian-Ukrainian war. Nowadays, this acute topic is in the spotlight of various language manipulations that cause numerous disinformation and profanity on social media platforms. The conducted experiment highlights three main stages of data annotation and underlines the main obstacles during machine annotation. Ultimately, we provide a fundamental statistical analysis of the obtained data, evaluation of models used for pseudo-labeling, and set further guidelines on how the scientists can leverage the corpus to execute more advanced research and extend the existing data samples without annotators' engagement. 

\end{abstract}

\section{Introduction}
The Russian invasion of Ukraine has been causing thousands of casualties, millions of displaced people, and severe economic and social consequences for many countries. The full-fledged escalation of the conflict broke out on February 24, 2022, when Russian forces trespassed the sovereign country's territory with flying jets and military vehicles \cite{ellyatt2022russian}. The wave of misinformation, panic, and mass hysteria took a toll on millions of Ukrainians during the first days of the invasion. Although the war has been ongoing for over a year, the problem of disinformation, misinformation, and harmful content identification across various social media platforms remains an open issue. The scarcity of well-annotated and verified warfare datasets is the main obstacle to developing high-quality models for offensive speech detection, disinformation, and misinformation classification \cite{poletto2021resources}.\par The study by \citet{Pierri_2023} is of particular interest as the authors examine Twitter accounts' creation and suspension dynamics based on tweets about the Russian-Ukrainian war. The scientists underline the vagueness of Twitter's policies regarding de-platforming. The most common soft-moderation tactics deployed by Twitter include down-ranking (lowering the visibility of certain content in users' feeds), "shadow banning" (hiding content from other users), and warning labels (tagging content as potentially harmful or inaccurate) \cite{papakyriakopoulos2022impact, ali2021understanding, pierri2022does}. As a result, some people fall victim to "shadow banning" by an algorithm's miscalculation. It might also apply to Ukrainian accounts that post messages in their native language but get down-ranked due to the incongruence of a moderation model. To implement a well-rounded neural network, one must have high-quality data, such as a labeled dataset for the Ukrainian language. Therefore, the study presents the first and only tagged Ukrainian corpus for offensive language detection in the context of the Russian-Ukrainian war.

The article's main objective is to describe a new data collection and labeling approach. The sensitive content of gathered tweets requires a rigorous algorithm to minimize the subjectivity of human evaluation. Hence a pseudo-labeling technique has been utilized at the second and third stages of the annotation. We also highlight the main challenges and limitations faced at different phases. In the end, some descriptive statistics and general analyses are offered to illustrate the potential and usefulness of the dataset for studying the offensive language in the context of the Russian-Ukrainian war. We hope this dataset will contribute to a better understanding of the offensive context in Ukrainian tweets and will be applied to various types of research on the level of the Russian dataset VoynaSlov and many English-annotated datasets \cite{chen2023tweets, park2022challenges}.

The paper is structured in the following way:
\begin{enumerate}
    \item Outline of the related works on the Russian-Ukrainian war together with available monolingual Ukrainian datasets;
    \item A closer look at the three stages of the data collection and annotation process, a description of the challenges that have occurred down the way;
    \item General data statistics of the obtained corpus and suggestions for further research works.
\end{enumerate}

\section{UA Corpus for Offensive Language Detection}

\subsection{Data collection}
5000 tweets were prior collected through an available Twitter streaming API service. In order to minimize the probability of acquiring tweets unrelated to the topic of war, we initially selected the ten most prominent and unique hashtags (Table~\ref{tab:hashtags}) which appeared at different periods of military actions in Ukraine. 

After the primary analysis of the social media platform, we can conclude that Ukrainians tend to write their tweets in English rather than Ukrainian. It can be addressed as an effort to attract more attention from Western countries to the situation in Ukraine. We also added a few general hashtags to the existing list (\#Ukraine, \#Russia, \#ukraine, \#russia) to get a more extensive collection of tweets. Other filters condition that a tweet should not be a reply or retweet, and the language of the content is strictly Ukrainian. The gathered messages cover a period from 09/2022 to 03/2023.

Although we explicitly stated the target language, some tweets were scrapped in Russian and Belarussian. Therefore, at this step, we eliminate duplicates (usually produced by bots), validate the language, and check a tweet’s correspondence to the war (it is necessary as we use a couple of general hashtags). Finally, only 2043 tweets remain in the dataset. 

\begin{table}[!h]
\resizebox{\columnwidth}{!}{%
\begin{tabular}{l}
\multicolumn{1}{c}{\textbf{Hashtags}}                       \\
\#RussiaIsATerroristState   \#russiaisaterroriststate      \\
\#WarInUkraine   \#warinukraine                            \\
\#Україна   \#українці                                     \\
\#BeBraveLikeUkraine   \#bebravelikeukraine \#braveukraine \\
\#UkraineWar   \#UkraineRussiaWar                          \\
\#StandWithUkraine                                         \\
\#ріквійни                                                 \\
\#Putin   \#путін                                          \\
\#СлаваУкраїні \#GloryToUkraine                            \\
\#FreeLeopards   \#freeleopards                           
\end{tabular}%
}
\renewcommand{\tablename}{Table}
\caption{Top ten unique hashtags related to the war in Ukraine.}
\label{tab:hashtags}
\end{table}

\subsection{Definitions of the offensive language}

Prior to the annotation stages we need to provide a comprehensive definition and criteria of what is considered as an instance of the offensive language use in our dataset. Every year a large amount of studies tackle the problem of offensive, abuse, hate and toxic language \cite{wiegand-etal-2021-implicitly-abusive, davidson2017automated, israeli-tsur-2022-free, saleem-etal-2022-enriching}. Despite the numerous studies that present an exhaustive outline and a definition of the offensive language, the scientists point out still occurring discrepancies between annotators \cite{sigurbergsson-derczynski-2020-offensive, goffredo-etal-2022-counter, ruitenbeek-etal-2022-zo, ross2017measuring}. We strive to minimize inconsistencies in inter-annotator agreement (IAA) by setting a clear-cut demarcation and regularities of what should be regarded as offensive content \cite{demus-etal-2022-comprehensive}. 

\citet{sigurbergsson-derczynski-2020-offensive} formulates the offensive language as a phenomenon that varies greatly and ranges from simple obscene language to more severe cases such as life threat, hate, bullying and toxicity. \citet{bretschneider2017detecting} states that hate speech, cyberhate and offensive language are umbrella terms used in the context of social media to denote offending or hostile message. Many researchers highlight that it remains hard to distinguish between offensive language and hate speech \cite{waseem2017understanding, sigurbergsson-derczynski-2020-offensive, waseem-hovy-2016-hateful, stamou-etal-2022-cleansing}. However, there exists some general agreement that hate speech is usually defined as "language that targets a group with the intent to be harmful or to cause social chaos" and can be identified as a subset of offensive language \cite{ sigurbergsson-derczynski-2020-offensive, schmidt-wiegand-2017-survey}. On the other hand, offensive language, is a broader category containing any type of profanity or insult \cite{ sigurbergsson-derczynski-2020-offensive}. As the UA corpus is a collection of annotated tweets gathered from the social media platform, we apply a definition provided by \citet{zampieri-etal-2019-predicting}, who determines that a message is offensive if it contains any form of foul language or a targeted offense, which can be stated implicitly or explicitly. The targeted offense may be insults, threats, and posts containing obscene language. 

\citet{zampieri2019semeval2019} introduces general guidelines for offensive language identification, its types and targets. \citet{waseem-hovy-2016-hateful} attempt to give the most rigorous criteria of what should be considered as an offensive message. The researchers highlight ten main points of any offensive tweet: "1) it uses a sexist or racial slur; 2) it attacks a minority; 3) it seeks to silence a minority; 4) it criticizes a minority (without a well founded argument); 5) it promotes, but does not directly use hate speech or violent crime; 6) it criticizes a minority and uses a straw man argument; 7) it blatantly misrepresents truth or seeks to distort views on a minority with unfounded claims; 8) it shows support of problematic hash tags; 9) it negatively stereotypes a minority; 10) it defends xenophobia or sexism; 11) it contains a screen name that is offensive". We add the direct citation from the article as it gives a thorough and concise summery of the Twitter’s rules and policies sections on abusive, violent and hateful behaviour. The tweets that contain any marked characteristics become suspended.  \footnote{\url{https://help.twitter.com/en/rules-and-policies/abusive-behavior}, \url{https://help.twitter.com/en/rules-and-policies/violent-speech},
\url{https://help.twitter.com/en/rules-and-policies/hateful-conduct-policy}}
We modify the criteria in the following way. A tweet is offensive if:
\begin{enumerate}
    \item it promotes xenophobia, uses sexist or racist slur;
    \item it implies the direct attack on a person or a group of people;
    \item it promotes violence or abuse (overtly through the profound language or covertly);
    \item it promotes misconception or misrepresentations that targets some violence or harm;
\end{enumerate}
We further utilize the defined points as the guidelines for annotators. 

\subsection{Stages of the annotation process}
There are three general scenarios for annotators selection: the subject-matter experts; individuals familiar with the subject background; and a crowdsourcing platform, where the annotators are only known after the process \cite{poletto2021resources}. Our major challenge during the recruiting period was the war context. Regardless of whether a person was inside Ukraine when the invasion started or outside – people perceive the atrocities of war similarly for many reasons: strong national identity, families or relatives that remain in Ukraine, etc. \cite{слюсаревський2022соціально} Nevertheless, we decided to assess the geographical location of annotators at the time of data processing to minimize the probability of biased opinions. We do not exclude those who reside in Ukraine. As a result, 15 people familiar with the topic of war agree to participate voluntarily in the rating procedure. Each of them is provided with guidelines on what should be evaluated as an offensive tweet. Among the sample, 8 participants reside outside Ukraine, and 7 – stay in Ukraine; 5 of them are professional linguists with some prior experience in data annotations, and others are academics from different fields. Women prevail over men in the sample (12 vs. 3). 

The whole annotation process is divided into three main iterations. The first stage includes 15 participants who manually annotate 300 tweets. Although the number of tweets is minimal, it is worth highlighting that the war is still ongoing, and new facts or crimes occur daily, which influence peoples' decisions. Moreover, the content of tweets is quite sensitive, which also impacts the general psychological sustainability of people to finish data annotation in one take. Considering the psychological factor, in the second stage, we strive to apply a pseudo-labeling technique \cite{arazo2020pseudo,kuligowska2021pseudo} to tag a batch of 700 tweets by fine-tuning RoBERTa \cite{minixhofer-etal-2022-wechsel} and ELECTRA \cite{stefan_schweter_2020_4267880} for the Ukrainian language using the Keras library \cite{gulli2017deep}. As the data is scarce, we obtain some inconsistent and biased labels; hence the three linguists who take part in the first stage and reside outside Ukraine are chosen to check the pseudo-labeled data. 

Consequently, we get both manual and machine annotation. Repeating the automatic tagging process for the remaining 1043 tweets, we gather a sample of uncertain messages (tweets whose probability lay between 0.40 to 0.55). Hence, the same three annotators adjudicate the pseudo-labels. 

Here we summarize the three iterations completed for the data annotation. Further, we provide more in-depth characteristics and results for every stage.

\newblock{\textbf{Results of the Stage I}}

The approach we acquire at the first stage of the data annotation is similar to the one described in the study by \citet{ruitenbeek-etal-2022-zo}. The researchers offer three criteria for labeling the data, where the first option states "EXPLICIT" if the content expresses profanity unambiguously on the lexical level. The message is "IMPLICIT" when it lacks the overt lexical markers of offensive language. The option "NOT" is chosen when no offense is found. Instead of the three-layer annotation approach, we offer the raters to choose between four categories: 
\begin{itemize}
    \item Offensive language, offensive sense;
    \item Neutral language, neutral sense;
    \item Offensive language, neutral sense;
    \item Neutral language, offensive sense
\end{itemize}
These labels allow people to make more accurate judgments about the context. Tweets under the labels "Offensive language, offensive sense" and "Neutral language, offensive sense" are straightforward in their semantic manifestation, which can be conveyed through explicit or implicit markers. On the other hand, tweets that fall under the categories "Offensive language, neutral sense" and "Neutral language, neutral sense" carry no offensive meaning but can be externalized through some harsh or inappropriate language. 
 
Fifteen people have completed the first iteration of data labeling. The number of participants appears to be significant; however, it has been agreed to keep a more extensive sample to achieve less biased results considering the nature of the material presented in the dataset. Selected annotators receive a link to the Google Form with a user-friendly interface and guidelines. 
 
\begin{figure}[!h]
\includegraphics[width=8cm]{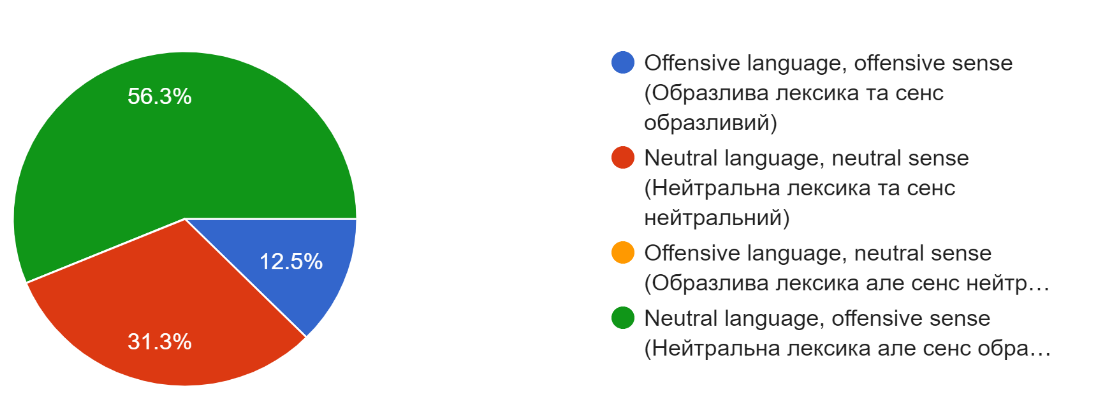}
\renewcommand{\figurename}{Figure}
\caption[english]{Explicitly neutral tweet that implies no offensive meaning.}
\centering
\label{figure:one}
\end{figure}

\begin{figure}[!h]
\includegraphics[width=8cm]{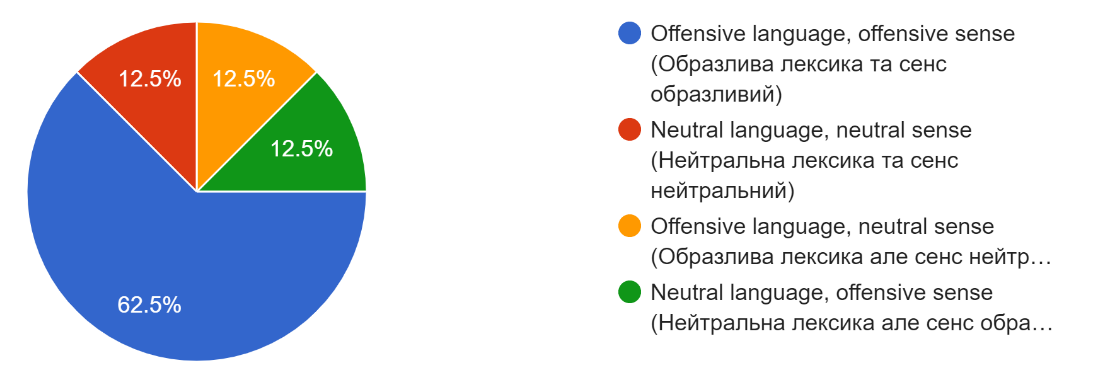}
\renewcommand{\figurename}{Figure}
\caption[english]{Explicitly offensive tweet with profanity words and overtly offensive meaning.}
\centering
\label{figure:two}
\end{figure}

When the annotation process is completed, we access the spreadsheet with answers and extract statistics for each tweet. Some examples are presented in Figure~\ref{figure:one} and Figure~\ref{figure:two}. Due to ethical policy we omit revealing the context of the tweets. 31\% of participants correctly identify that the first tweet carries a neutral sense, whereas 25\% have stated that the message from Figure~\ref{figure:two} has no offense.\footnote{In out survey neutral or offensive "sense" equals neutral or offensive "meaning" of a tweet. These notions are used interchangeably here.}

Before proceeding to the second stage of the annotation procedure, we measure the IAA (the inter-annotator agreement) to evaluate the general quality of the acquired labels \cite{ artstein2017inter}. Since 15 participants took part in the labeling process, we used the Fleiss Kappa score to assess the IAA \cite{fleiss1971measuring}. Cohen’s Kappa is useful if the number of annotators is no more than 2. Hence it is not applicable in our case \cite{cohen1960coefficient}, while Krippendorff’s alpha is more relevant for collections with some missing values \cite{krippendorff1980validity}. We collapse four categories into offensive and neutral based on the tweet’s meaning and utilize an open-source statistical Python library to calculate the Fleiss Kappa\footnote{\url{ https://www.statsmodels.org/dev/generated/statsmodels.stats.inter_rater.fleiss_kappa.html}}.The inter-annotator agreement score at this stage is 0.384, indicating fair agreement between raters.

\newblock
\begin{table*}[!h]
\centering
\scalebox{1.0}{%
\begin{tabular}{lccc}
\hline
\multicolumn{4}{c}{\textbf{Results after the Stage I}}            \\ \hline
\multicolumn{1}{l|}{\textbf{Model}} & \multicolumn{1}{l}{\textbf{Recall OF}} & \multicolumn{1}{l}{\textbf{Recall NON-OFF}} & \multicolumn{1}{l}{\textbf{F1 Macro}} \\ \hline
\multicolumn{1}{l|}{DistilBERT   (multilingual)}  & .20 & .95 & .35 \\ \hline
\multicolumn{1}{l|}{RoBERTa   + BiLSTM}           & .86 & .62 & .65 \\ \hline
\multicolumn{1}{l|}{ELECTRA   + ReLU Dense layer} & .83 & .63 & .73 \\ \hline
\multicolumn{1}{l|}{ELECTRA   + BiLSTM}           & .71 & .82 & .69 \\ \hline
\multicolumn{4}{l}{\textbf{Results after the Stage II}}             \\ \hline
\multicolumn{1}{l|}{RoBERTa   + BiLSTM}           & .59 & .89 & .65 \\ \hline
\multicolumn{1}{l|}{ELECTRA   + ReLU Dense layer} & .82 & .78 & .76 \\ \hline
\multicolumn{1}{l|}{ELECTRA   + BiLSTM}           & .74 & .80 & .70 \\ \hline
\multicolumn{4}{l}{\textbf{Results after the Stage III}}            \\ \hline
\multicolumn{1}{l|}{RoBERTa   + BiLSTM}           & .60 & .95 & .69 \\ \hline
\multicolumn{1}{l|}{ELECTRA   + ReLU Dense layer} & .73 & .90 & .72 \\ \hline
\multicolumn{1}{l|}{ELECTRA   + BiLSTM}           & .66 & .96 & .74 \\ \hline
\end{tabular}%
}
\renewcommand{\tablename}{Table}
\caption{Results after each stage.}
\label{tab:stages-res}
\end{table*}

In the second iteration, we aim to utilize a pseudo-labeling technique for data annotation. This approach has demonstrated rigorous and consistent results in the computer vision domain \cite{ iscen2019label, xie2020self} and recently gained much attention in NLP research \cite{ahmed2011pseudo, li2018pseudo}. We follow a methodology offered by \citet{ kuligowska2021pseudo}, where authors use the DistilBERT model to distinguish questions from answers. 

The 300 manually annotated tweets are applied for fine-tuning four neural network architectures:
\begin{enumerate}
    \item DistilBERT (multilingual) (baseline model)
    \item RoBERTa + BiLSTM
    \item ELECTRA + ReLU Dense layer
    \item ELECTRA + BiLSTM
\end{enumerate}

An exhaustive list of the pre-processing steps and hyperparameters is provided in Appendix ~\ref{sec:appendixa} for replicability. DistilBERT is trained for 104 languages, so we do not expect it to perform well for this particular task.\footnote{\url{ https://huggingface.co/distilbert-base-multilingual-cased }} On the other hand, we apply a specific type of ELECTRA model (discriminator) trained solely on the Ukrainian data specifically for the text classification tasks \footnote{\url{https://huggingface.co/lang-uk/electra-base-ukrainian-cased-discriminator }}, anticipating it outperforms other architectures.  

The models are trained on the train split (80\%) and evaluated against the non-overlapping test split (20\%). At this stage, the architectures are compared using the Recall scores of two classes. The choice of this metric is driven by the imbalance of the data, where 60\% of the annotated tweets belong to the neutral class; hence the models are prone to overfit. The Recall score gives insight into the sensitivity or true positive rate prioritized at this stage. We utilize the F1 score for the final iteration as the number of samples increases. Besides, we opt for a more general statistical evaluation provided by the evaluation metric that measures the model's accuracy. At this stage, the primary objective is to collect more or less solid probabilities for each tweet. Table~\ref{tab:stages-res} presents the Recall scores for each model.

Due to the scarcity of data, three out of four models result in overfitting. ELECTRA + BiLSTM architecture has shown a more rigorous outcome compared to others. Nevertheless, we are unsure about the probabilities assigned for the unlabelled 700 tweets. Therefore, two raters from the previous sample of 15 people are chosen based on their professional training in linguistics and the 0.625 pairwise Cohen's Kappa score after the first iteration \cite{landis1977application}, which indicates substantial agreement. The paper's corresponding author serves as an adjudicator of the final label.

A batch of 700 tweets has been pre-annotated by the chosen model architecture and verified by three annotators. The second iteration results in a set of 1000 annotated and justified tweets.

\newblock
\textbf{Results of the Stage III}

We prune the baseline model and evaluate the three transformer models using the same train/test split on the labeled 1000 tweets. Table~\ref{tab:3iter} displays each architecture's Recall and F1 (on the offensive) scores.

We utilize a simple ELECTRA + ReLU Dense layer model at this stage as it slightly outperforms the previous ELECTRA architecture. We also strive to minimize the overfitting of the BiLSTM layer at the previous stage. Correspondingly we get the probabilities for the 1043 unlabeled tweets. All tweets in the range of [0.40, 0.55] have been submitted for manual verification by the same three raters.

Subsequently, we have obtained the total annotated corpus of 2043 tweets partially labeled by the selected individuals and partially by the neural networks. The final Fleiss Kappa remains close to the one obtained in the second stage - 0.814. 

The Table~\ref{tab:stages-res} describes the evaluation of the three neural networks on the annotated dataset. As we can conclude, the performance of each model has improved; models demonstrate robust and consistent results regardless of the number of performed iterations.

\begin{figure}[!h]
\includegraphics[width=8cm]{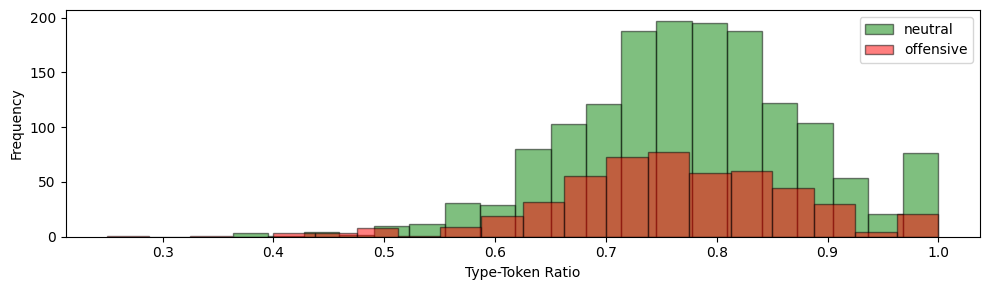}
\renewcommand{\figurename}{Figure}
\caption[english]{A distribution of type-token ratio in neutral and offensive subsets of the dataset.}
\centering
\label{figure:TTR}
\end{figure}

\begin{figure}[!h]
\includegraphics[width=8cm]{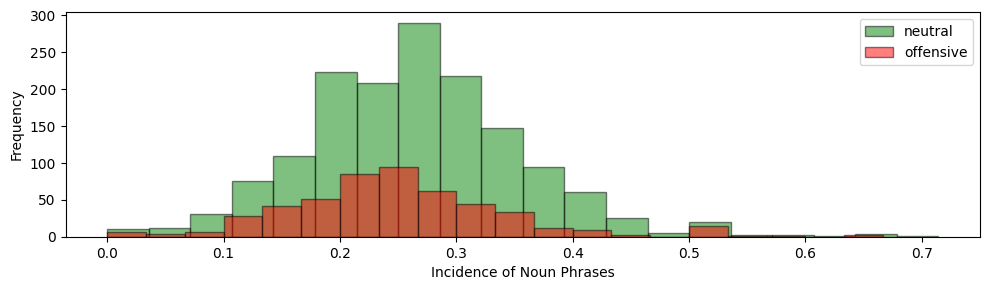}
\renewcommand{\figurename}{Figure}
\caption[english]{The 25 most frequent words in the subset of offensive tweets (Ukrainian language).}
\centering
\label{figure:NPS}
\end{figure}

\section{Data Statistics}
The corpus is available as a public GitHub\footnote{The link will be added after the blind review.} repository, enabling further research on offensive language detection and war rhetoric in the Ukrainian language. As Twitter’s Terms \& Conditions\footnote{\url{https://developer.twitter.com/en/docs/twitter-api/compliance}} prohibit any public release of texts or metadata of tweets, we provide tweets’ IDs and labels (1 – offensive; 0 – neutral). Scientists can use third-party tools such as Hydrator\footnote{\url{https://github.com/DocNow/hydrator}} or Twarc\footnote{\url{https://github.com/DocNow/twarc}}to obtain the raw context. 

The UA Corpus for Offensive Language Detection in the Context of the Russian-Ukrainian War incorporates 500 offensive tweets and 1543 neutral gathered from 1020 unique users. The Table~\ref{tab:25neutral} offers the English translation of the 25 most frequent words in the subset of neutral tweets. Subsequently, the Table~\ref{tab:25off} lists the 25 most frequent words from the offensive subset. The corresponding pie charts with the Ukrainian equivalents and word-clouds can be found in Appendix ~\ref{sec:appendixb}.

We can conclude that the term "the Armed Forces of Ukraine" is equally significant for offensive and neutral tweets. "Ukraine" is used more frequently in the neutral context rather than offensive. The word "Russia" dominates in the offensive tweets and ranks less for neutral. Noticeable that the top five words of the offensive tweets do not incorporate any obscene or profane language. 

\begin{table}[!h]
\centering
\scalebox{0.7}{%
\begin{tabular}{|c|c|}
\hline
\textbf{Word}                         & \textbf{Statistics}  \\
Ukraine   (different declinations)                                & 9.4\%,   8.5\%, 8.2\%, 3.2\% \\
The Armed Forces of Ukraine [ЗСУ]                        & 8.0\%                        \\
glory                                 & 6.1\%                \\
war   (different declinations)        & 5.1\%   , 4.1, 2.4\% \\
RF, Russia (different declinations) & 4.3\%,   3.8\%, 3.5\%        \\
people                                & 4.1\%                \\
day                                   & 3.5\%                \\
life                                  & 3.1\%                \\
Ukrainians   (different declinations) & 3.1\%,   2.8\%,      \\
the   USA                             & 2.8\%                \\
anti-aircraft warfare               & 2.6\%                \\
NATO                                  & 2.5\%                \\
rockets                               & 2.3\%                \\
region                                & 2.3\%                \\
victory                               & 2.1\%                \\
country                               & 2.1\%                \\ \hline
\end{tabular}%
}
\renewcommand{\tablename}{Table}
\caption{The English translation of the 25 most frequent words among the neutral tweets.}
\label{tab:25neutral}
\end{table}

\begin{table}[!h]
\centering
\scalebox{0.6}{%
\begin{tabular}{|c|c|}
\hline
\textbf{Word}                                     & \textbf{Statistics}                        \\
Russians (different declinations)               & 9.9\%,   5.1\%, 4.3\%, 4.1\%, 3.7\%, 2.8\% \\
The Armed Forces of   Ukraine [ЗСУ]    & 6.1\%                                      \\
Ukraine (different declinations)                & 5.3\%,   4.7\%, 3.9\%                      \\
war                  & 5.1\%                                      \\
country                                           & 4.3\%                                      \\
people                                            & 4.1\%                                      \\
what                                              & 4.1\%                                      \\
glory                                             & 3.9\%                                      \\
go f**k yourself                                  & 3.7\%                                      \\
rockets                                           & 3.4\%                                      \\
want                                              & 3.4\%                                      \\
Putin                                             & 3.2\%                                      \\
fag***s                                           & 3.0\%                                      \\
dumb                                              & 3.0\%                                      \\
hate (verb)                                       & 2.8\%                                      \\
day                                               & 2.8\%                                      \\ \hline
\end{tabular}%
}
\renewcommand{\tablename}{Table}
\caption{The English translation of the 25 most frequent words among the offensive tweets.}
\label{tab:25off}
\end{table}
Moreover, we apply an open-source tool for corpus analysis - the StyloMetrix\footnote{\url{https://github.com/ZILiAT-NASK/StyloMetrix\#readme}} to extract grammatical features that help to set a boundary between the offensive vs. the neutral language of tweets \cite{okulskastyles}. For instance, the Figure~\ref{figure:TTR} indicates an aggregated type-token ratio of tweets. Even though the neutral tweets dominate in the dataset, their overall frequency is higher; we can still trace the tendency of offensive tweets to be slightly shorter than the neutral ones. Another example is the incidence of noun phrases (Figure~\ref{figure:NPS}). The mean value of NPs in the offensive subset is close to 0.25, whereas the mean of neutral NPs shifts closer to 0.3.   
\section{Related Works}
\subsection{Existing datasets on the Russian-Ukrainian war}
Since the beginning of the full-scale Russian invasion, many corpora related to the war have been produced to research disinformation, warfare, misinformation, and political discourse. The datasets described in this section primarily aim to provide essential statistical evaluations of the texts related to the war.\footnote{\url{https://conflictmisinfo.org/datasets/}} The existing warfare corpora can be divided into two broad categories: multilingual and monolingual, mainly collected via Twitter's streaming API\footnote{\url{https://developer.twitter.com/en/docs/tutorials/consuming-streaming-data}} or other web-scraping tools. For instance, a dataset by \citet{chen2023tweets} incorporates over 570 million tweets in more than 15 languages. The researchers offer a concise overview of their corpus's top languages and keywords. Another publicly available multilingual dataset is "UKRUWAR22: A collection of Ukraine-Russia war related tweets," \cite{g0me-wa71-22} comprising 55186 unique tweets in 57 languages. "Twitter dataset on the Russo-Ukrainian war"\cite{shevtsov2022twitter} is a web-based analytical platform that daily updates the analysis of the volume of suspended/deactivated accounts, popular hashtags, languages, and positive/negative sentiment of tweets. A similar corpus by \citet{haq2022twitter} includes 1.6 million tweets; while the project is ongoing, it outlines the keywords assessment and language diversity. The listed multilingual datasets contain a profound amount of raw data that can be used to make broad statistical inferences about cross-language and sentiment analysis or as a tool for unsupervised data mining for topic detection, author identification, disinformation, and misinformation pattern extraction.

On the other hand, the monolingual corpora on the Russian-Ukrainian war essentially cover English sources and are significantly underrepresented for other languages. The online English dataset by the Social Media Labs\footnote{\url{https://conflictmisinfo.org/}} gives a deeper insight into an alleged chemical attack in Mariupol. The focus was to construct the retweet network and to identify Ukraine's seven most retweeted accounts that broadcasted this topic. The corpus by \citet{fung2022weibo} is a collection of over 3.5M user posts and comments in Chinese from the popular social platform Weibo. The gathered data can be a rich resource for propaganda and disinformation analysis in China. Another group of researchers created the Twitter dataset, which encompasses only original tweets in the English language, excluding retweets or quotes \cite{pohl2022twitter}. The data covers one weeks before the war and one week after the onset of the Russian invasion. "VoynaSlov" is a corpus that contains only the Russian language texts scraped from Twitter and a Russian social platform VKontakte \cite{park2022challenges}. The dataset includes 38M posts subdivided into two groups: state-affiliated texts and notes from independent Russian media outlets. The researchers state that the main objective is to use the obtained data to capture Russian government-backed information manipulation, which can be regarded as disinformation and propaganda.

Despite the plethora of datasets, most lack validation criteria that validate that the gathered texts are related to the topic of war. We assess this drawback while creating a well-grounded monolingual Ukrainian dataset. Moreover, the researchers highlight that the data scraped through the Twitter streaming API is not entirely random, which may result in some biases \cite{shevtsov2022twitter, pohl2022twitter}. Unfortunately, we cannot escape this shortcoming as the presented dataset is collected through Twitter's API.  
\section{Conclusions and Future Work}

The study introduces the first Ukrainian dataset for offensive language detection in the context of the Russian-Ukrainian war. We propose a new method for annotating sensitive data using a pseudo-labeling algorithm with transformer models and human validation. In the first iteration, the annotators choose between four labels that capture tweets' explicit and implicit offensive meaning. Then, the four labels are merged into two categories: offensive and neutral, depending on the context. We apply three main neural network architectures and obtain satisfactory results in the following two stages of data collection. The best-performing architecture in the second stage is ELECTRA + BiLSTM; however, it tends to overfit due to the small corpus size, which consists of only 300 tweets. Therefore, we submit 700 automatically annotated tweets for verification to three annotators. In the last stage, we collect the logits from ELECTRA + ReLU Dense layer architecture. If the tweet's probability falls within [0.40, 0.55], its label is adjudicated by the raters. The final corpus comprises 500 offensive tweets and 1543 neutral tweets collected from 1020 unique users.

We present the descriptive statistics of the collected data by extracting the 25 most frequent words from each class and using the StyloMetrix tool to identify some grammatical features that differentiate offensive language from neutral language.

In future work, we plan to enlarge and balance the dataset and develop more robust neural networks for offensive language detection in Ukrainian. We also aim to apply the established criteria and terminology of offensive language to create a general Ukrainian multilabel dataset for abusive and hate speech detection.

\section*{Limitations}
The dataset has a few limitations worth noticing:
\begin{enumerate}
    \item \textbf{A human factor.} The collected tweets present the warfare content, and as the war is ongoing, people carry some bias, prejudice, and emotions that can influence any judgment, even professionally trained annotators. Hence, there remains room for bias in the obtained labels. 
    \item \textbf{Data labeling.} One can argue that hashtags and emojis play a significant role in data labeling. However, we eliminated them by explicitly mentioning to annotators not to consider them. We adhere to this rule because of the tweets' context. If people were to consider the hashtags, their opinion would have fluctuated even more, and in the end, we would not have achieved any rigorous and agreed annotation. 
    \item \textbf{Twitter API access.} In compliance with Twitter’s rules and content-sharing policies\footnote{\url{https://help.twitter.com/en/rules-and-policies\#twitter-rules}}, we must provide only tweet IDs and labels, which can lead to data loss in further dehydration because some accounts can be suspended or banned at the time of content extraction. Besides, the Twitter stream rate limit may restrict some content during the data scraping process, consequently bringing that bias to the corpus.
    \item \textbf{Imbalance of data.} The number of neutral tweets is dominant in the dataset, which can cause incongruence in neural network models and limit their performance. We plan to balance the dataset during the following stages of its development.
\end{enumerate}

\section*{Ethics Statement}
Scientists who use this dataset need to understand the sensitivity of the context they aim to research. The inferences, conclusions, and statements they can make based on the content of the tweets may have a powerful influence on many people and their opinion on this war. Hence, the researchers need to be objective and rational while delivering their work. Moreover, one has to remember that collected tweets present only a small subset of the Twitter’s data. Therefore, the bias and limitations have to be explicitly stated in their work. 

Furthermore, we provide the content of tweets, excluding accounts’ IDs, retweets, links, or any personal information, only upon explicit request and specifically to scientists for academic purposes. The academics granted the access should comply with our main conditions to not redistribute the corpus to third parties and not publish it as an open-source. Therefore, we satisfy Twitter's regulations on this issue. 

\bibliography{anthology,custom}
\bibliographystyle{acl_natbib}

\appendix
\section{Appendix A: Guidelines for reproducibility.}   
\label{sec:appendixa}

The data cleaning and pre-processing for the second and third iterations:
\begin{enumerate}
    \item lowercasing of all words
    \item all users’ mentions were eliminated
    \item all URLs were deleted
    \item emojis were excluded
    \item hashtags with the following were removed
    \item extra blank spaces were replaced with a single space
    \item extra blank new lines were removed
\end{enumerate}

Models’ hyperparameters:
\begin{itemize}
    \item \textbf{DistilBERT (multilingual)}

\textbf{Layers:}

trf\_model(input\_word\_ids, attention\_mask)

Flatten layer

Dense layer (1, activation= 'sigmoid')

\textbf{Hyperparameters:}
epochs = 4; batch size = 32; Adam optimizer with learning rate = 5e-5.

    \item \textbf{RoBERTa + BiLSTM}

\textbf{Layers:}

trf\_model(input\_word\_ids, attention\_mask)

SpatialDropout1D(0.3)

Bidirectional(LSTM(128, return\_sequences=True))

Dropout(0.5)

Bidirectional(LSTM(128, return\_sequences=True))

Dense(128, activation='relu')

Dense(1, activation='sigmoid')

\textbf{Hyperparameters:}
epochs = 3; batch size = 16; Adam optimizer with learning rate = 3e-5.

    \item \textbf{ELECTRA + Dense layer with ReLU activation}

\textbf{Layers:}

trf\_model(input\_word\_ids, attention\_mask)

Flatten layer

Dense(128, activation="relu")

Dense layer (1, activation= 'sigmoid')

\textbf{Hyperparameters:}
epochs = 3; batch size = 16; Adam optimizer with learning rate = 3e-5.

    \item \textbf{ELECTRA + BiLSTM}

\textbf{Layers:}

trf\_model(input\_word\_ids, attention\_mask)

SpatialDropout1D(0.3)

Bidirectional(LSTM(128,return\_sequences=True))

Dropout(0.5)

Bidirectional(LSTM(128,return\_sequences=True))

Dense(128, activation='relu')

Dense(1, activation='sigmoid')

\textbf{Hyperparameters:}
epochs = 3; batch size = 16; Adam optimizer with learning rate = 2e-5.
\end{itemize}

\appendix
\section{Appendix B: General statistics of the dataset.}
\label{sec:appendixb}

\begin{figure}[!h]
\centering
\includegraphics[width=6cm]{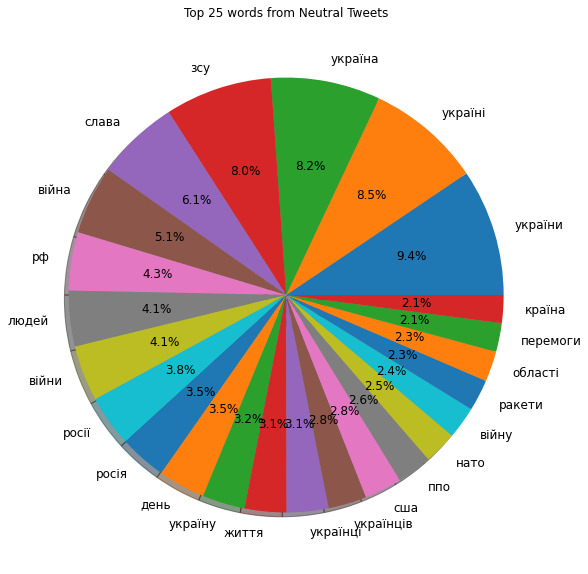}
\renewcommand{\figurename}{Figure}
\caption[english]{The 25 most frequent words in the subset of neutral tweets (statistics).}
\centering
\label{figure:neutral}
\end{figure}

\begin{figure}[!h]
\centering
\includegraphics[width=6cm]{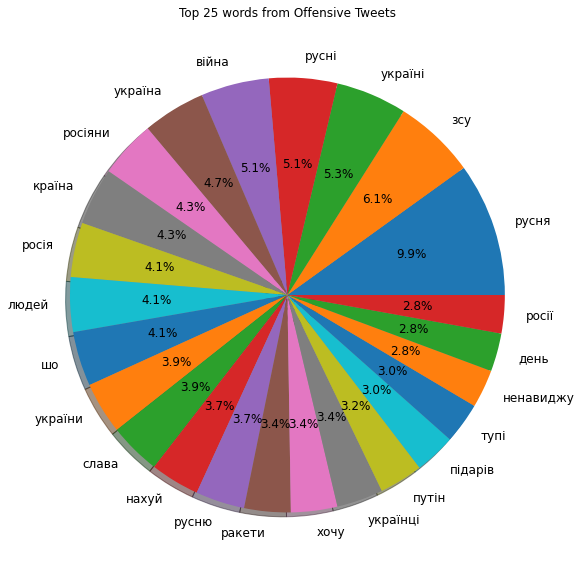}
\renewcommand{\figurename}{Figure}
\caption[english]{The 25 most frequent words in the subset of offensive tweets (statistics).}
\centering
\label{figure:neutral}
\end{figure}

\end{document}